%% file: reference.tex
\documentclass{article}

\PassOptionsToPackage{numbers, sort&compress}{natbib}

\usepackage{arxiv}

\usepackage[utf8]{inputenc} 
\usepackage[T1]{fontenc}    
\usepackage{hyperref}       
\usepackage{url}            
\usepackage{booktabs}       
\usepackage{amsfonts}       
\usepackage{nicefrac}       
\usepackage{microtype}      
\usepackage{xcolor}         

\usepackage{amsmath}
\usepackage{mathtools}

\usepackage{graphicx}
\usepackage{float}
\usepackage{stmaryrd}
\usepackage{subcaption}
\usepackage[export]{adjustbox}
\usepackage{xcolor,pifont}
\newcommand{\cmark}{\ding{51}}%
\usepackage{multirow}
\usepackage{booktabs}
\usepackage{pdfpages}
\usepackage{wrapfig}

\usepackage{caption}

\usepackage{soul}
\usepackage[textsize=tiny]{todonotes}

\DeclareUnicodeCharacter{2212}{-}

\title{libcll: an Extendable Python Toolkit for Complementary-Label Learning}

\author{
    Nai-Xuan Ye,\:
    Tan-Ha Mai,\:
    Hsiu-Hsuan Wang,\:
    Wei-I Lin,\:
    Hsuan-Tien Lin \\
    National Taiwan University \\
    \texttt{\{b09902008, d10922024, b09902033, r10922076, htlin\}@csie.ntu.edu.tw}
}
\date{}

\begin{document}

\maketitle

\begin{abstract}
\input{Sections/Abstract}
\end{abstract}

\section{Introduction}
\input{Sections/Introduction}

\section{Preliminaries and related works}
\input{Sections/Problem}

\section{Library design}
\input{Sections/Design}

\section{Benchmark experiments}
\input{Sections/Experiment}
\section{Conclusion}\label{sec:conclusion}
\input{Sections/Conclusion}

\section{Limitations and Future Work}\label{sec:limitation}
\input{Sections/Limitation}

\clearpage
\bibliography{reference.bib}
\bibliographystyle{unsrt}

\clearpage

\end{document}

%% file: Sections/Abstract.tex
Complementary-label learning (CLL) is a weakly supervised learning paradigm for multiclass classification, where only complementary labels---indicating classes an instance does not belong to---are provided to the learning algorithm. Despite CLL's increasing popularity, previous studies highlight two main challenges: (1) inconsistent results arising from varied assumptions on complementary label generation, and (2) high barriers to entry due to the lack of a standardized evaluation platform across datasets and algorithms. To address these challenges, we introduce \texttt{libcll}, an extensible Python toolkit for CLL research. \texttt{libcll} provides a universal interface that supports a wide range of generation assumptions, both synthetic and real-world datasets, and key CLL algorithms. The toolkit is designed to mitigate inconsistencies and streamline the research process, with easy installation, comprehensive usage guides, and quickstart tutorials that facilitate efficient adoption and implementation of CLL techniques. Extensive ablation studies conducted with \texttt{libcll} demonstrate its utility in generating valuable insights to advance future CLL research.

%% file: Sections/Introduction.tex

In many real-world applications, training effective classifiers typically depends on obtaining high-quality, accurate labels. However, acquiring such labels is often difficult and costly. To address this challenge, many researchers have turned their attention to weakly supervised learning (WSL), a methodology aimed at training reliable classifiers using only incomplete, imprecise, or inaccurate data~\cite{sugiyama2022machine, zhou2018brief}. Numerous WSL studies have been conducted to extend our understanding of machine learning capabilities, covering topics such as complementary labels~\cite{ishida2017learning, scl2020}, multiple complementary labels~\cite{mul-comp1, mcl2020}, noisy labels~\cite{noisy_label}, and learning from partial labels~\cite{partial_labels}.

This work focuses on complementary-label learning (CLL), a WSL problem where each label indicates only a class to which a data instance \textit{does not belong}~\cite{ishida2017learning}. CLL aims to train models with these complementary labels while still enabling accurate predictions of the ordinary labels during testing. CLL makes machine learning more practical in scenarios where obtaining ordinary labels is difficult or costly~\cite{ishida2017learning}. Additionally, CLL broadens our understanding of machine learning's practical potential under limited supervision.

Current research on CLL has introduced numerous learning algorithms~\cite{scl2020, fwd2018, gao2021discriminative, cpe2023} that have been evaluated using a diverse range of datasets, from synthetic datasets based on varied complementary-label generation assumptions to real-world datasets~\cite{clcifar2023}.
However, the performance of these algorithms often varies significantly across studies due to differences in underlying label-generation assumptions, the absence of a standardized evaluation platform, and the use of diverse network architectures~\cite{scl2020, fwd2018, ishida2017learning, cpe2023}. Establishing a fair, reproducible, and stable evaluation environment is therefore essential for advancing CLL research. For instance, variations in network architectures, such as the use of ResNet18~\cite{xu2019generativediscriminative, clcifar2023} versus ResNet34~\cite{fwd2018, scl2020}, contribute to inconsistencies in performance and hinder fair comparisons across studies. Furthermore, most CLL research has not publicly released implementations~\cite{mcl2020, cpe2023, scl2020, ComCo2023}, particularly regarding details like loss calculation and data pre-processing. This lack of accessibility presents a challenge for researchers seeking to validate and build upon existing work in CLL.

\begin{figure}[htb]
\centering
    \includegraphics[width=0.90\linewidth]{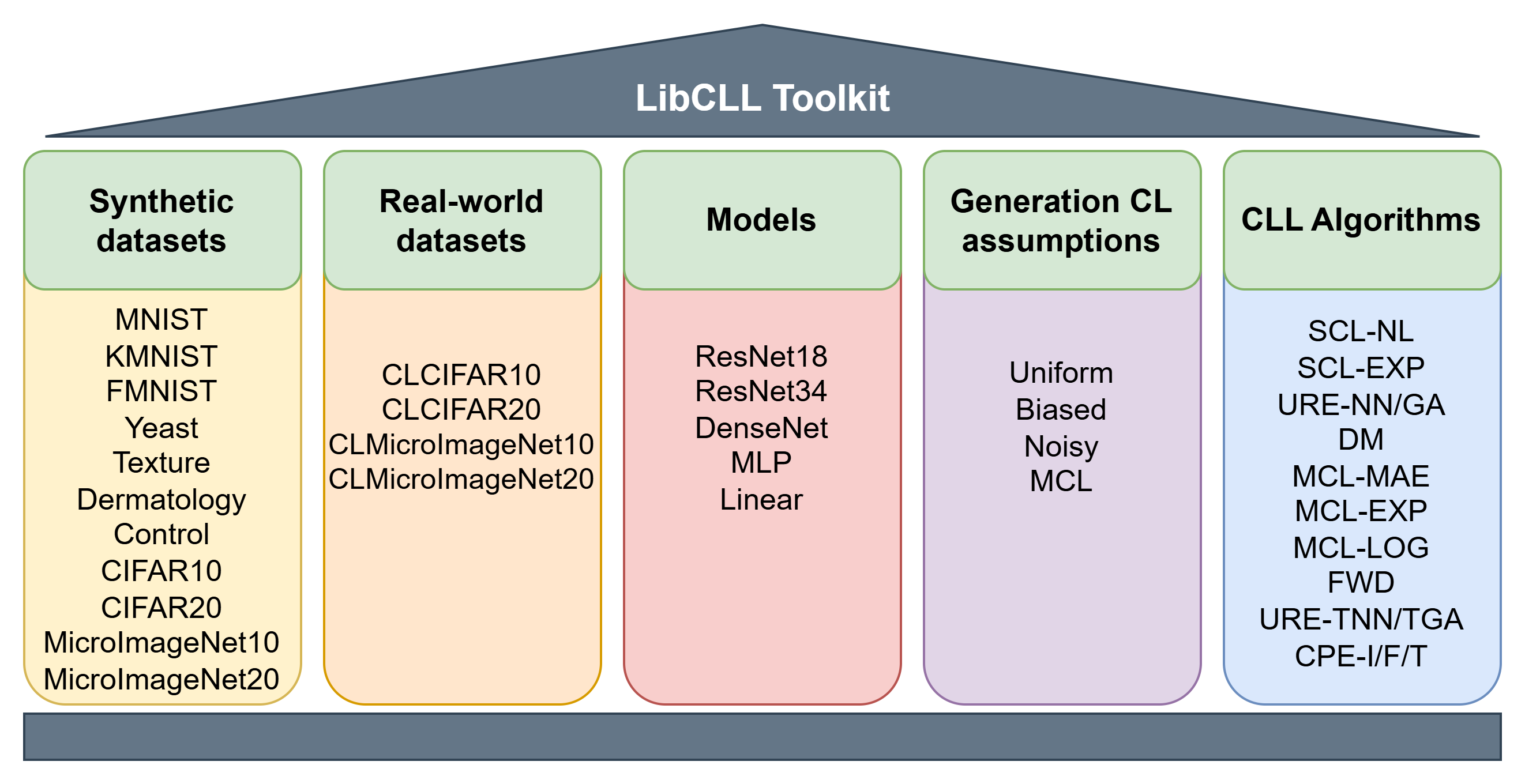}
    \caption{Coverage of the \texttt{libcll} Toolkit: This figure provides an overview of the key components included in the \texttt{libcll} toolkit, which encompasses 15 datasets spanning synthetic and real-world scenarios, 5 commonly used models in Complementary Label Learning (CLL), 4 CLL assumptions, and 14 CLL algorithms. To the best of our knowledge, \texttt{libcll} is the first comprehensive toolkit specifically dedicated to CLL.}
    \label{libcll_coverage}
\end{figure}
To enable meaningful comparisons among CLL algorithms and create a user-friendly environment for implementation and innovation, we introduce \texttt{libcll}, a complementary-label learning toolkit built with PyTorch-Lightning. This toolkit standardizes the evaluation process while offering extensive customization options, making it easier for researchers to develop, test, reproduce, and refine algorithms. By performing comprehensive benchmark experiments across established CLL datasets, various algorithms, and a range of complementary-label distributions, as illustrated in Figure \ref{libcll_coverage}, 
\texttt{libcll} provides a robust and reproducible evaluation framework. Our goal is for \texttt{libcll} to accelerate progress in CLL research and foster a collaborative research community.
Furthermore, \texttt{libcll} includes functions to generate complementary labels using a user-defined transition matrix and supports multiple complementary-label generation methods. Additionally, \texttt{libcll} offers extensive benchmarks across a wide range of datasets, from synthetic to real-world, using various CLL algorithms and complementary label distributions. Figure \ref{libcll_coverage} illustrates the comprehensive coverage of \texttt{libcll}. This toolkit, together with these benchmark results, provides the community with a holistic view of CLL, allowing researchers to assess the strengths and weaknesses of different approaches, pinpoint areas for improvement, and develop more resilient algorithms. By standardizing evaluation metrics and experimental setups, \texttt{libcll} promotes reproducibility and comparability, driving progress in the field and encouraging collaborative research efforts.

In summary, the contributions of our \texttt{libcll} toolkit are as follows: \begin{itemize} 
\item To the best of our knowledge, \texttt{libcll} is the first publicly available toolkit for CLL, now accessible at \url{https://github.com/ntucllab/libcll}. 
\item We introduce \texttt{libcll}, a platform that offers in-depth insights into Complementary Label Learning (CLL) by benchmarking 15 datasets, 14 algorithms, and diverse label distributions, while supporting customization of CLL components and highlighting each approach's strengths and limitations (see Figure \ref{libcll_coverage}).
\end{itemize}

%% file: Sections/Problem.tex
\subsection{Complementary-Label Learning}
In ordinary multi-class classification, a dataset $D = \{(x_i , y_i)\}^N_{i=1}$ is provided to the learning algorithm, where $N$ denotes the number of samples, and the dataset is i.i.d. sampled from an unknown distribution.
For each $i$, $x_i \in \mathbb{R}^d$ represents the $d$-dimension feature of the $i^{th}$ sample and $y_i \in \mathbb{R}^K$, where $K$ denotes the total number of classes in the dataset, with $K > 2$. 
The set $[K] = \{1, 2, . . . , K\}$ represents the possible classes to which  $x_i$ can belongs. 
The objective of the learning algorithm is to learn a classifier $f(x): \mathbb{R}^d \rightarrow \mathbb{R}^K$ that minimizes the classification risk: $E_{(x, y) \sim D}[\ell(f(x), e_y)]$, where $\ell$ is the loss function and $e_y$ is one-hot vector corresponding to label $y$. 
The predicted label $\hat y$ for an instance $x$ is determined by applying the argmax function on $f(x)$, i.e. $\hat y = \mathop{\arg\max}_{k \in K} f(x)_k$, where $f(x)_k$ represents the $k^{th}$ output component of $f(x)$.

In contrast to ordinary-label learning, complementary-label learning (CLL) shares the same goal of training a classifier but learns from a different label set. 
In CLL, the ground-truth label $y$ is not accessible to the learning algorithm. 
Instead, a complementary label $\bar y$ is provided, which is a class that the instance $x$ does not belong to. The training set of complementary becomes $\bar{D} = X \times \bar{Y}$, where $X$ is input space and $\bar{Y}$ is the corresponding complementary labels space.
The objective of CLL remains consistent with that of ordinary multi-class classification: to learn a classifier capable of predicting the correct label of unseen instances, even when trained on a complementary dataset $\bar D = \{(x_i , \bar y_i)\}^N_{i=1}$.
\subsection{Assumptions of complementary labels}

Complementary-label learning aims to develop a classifier under the guidance of weak supervision from complementary labels. For synthetic data generation, prior studies assume the distribution of complementary labels only depends on the ordinary labels instead of features; thus, $P(\bar y \mid x, y) = P(\bar y \mid y)$. The transition probability $P(\bar y \mid y)$ is often represented by a $K \times K$ transition matrix $T$, with $T_{ij} = P(\bar y=j \mid y=i)$.

The transition matrix can be further classified into three categories: \begin{itemize} 
\item \emph{Uniform} \cite{ishida2017learning}: All complementary labels are uniformly and randomly selected from $K-1$ classes. Based on this assumption, the transition matrix is $T = \frac{1}{K-1}(1_K - I_K)$. 
\item \emph{Biased} \cite{fwd2018}: Any transition matrix that is not uniform is biased. 
\item \emph{Noisy} \cite{cpe2023}: A portion of the true labels are mislabeled as complementary labels. Thus, the diagonals of the transition matrix are not necessarily zero. 
\item \emph{MCL} \cite{mcl2020}: Each instance has more than one complementary label.
\end{itemize}

Additionally, due to the difficulty of obtaining a transition matrix in real-world scenarios, CONU \cite{conu2023} proposed \emph{SCAR} (Selected Completely At Random), where the generation of complementary labels is independent of both instances' features and ground-truth labels; that is, $P(\bar y \mid x, y) = P(\bar y) = c_k$, where $c_k$ is a constant related only to the $k$-th class.
Furthermore, some studies, such as MCL \cite{mcl2020}, assume that each instance can have multiple complementary labels. This approach involves randomly generating a label set $\hat y$ where $1 \leq [\hat y] \leq K - 1$ and then asking annotators whether the given label set $\hat y$ contains the true label.

\input{Tables/cll-assumptions}

The uniform assumption for complementary labels originates from a label-collection perspective. In this approach, each data instance is associated with a randomly assigned label, and workers verify the validity of each label by responding with either 'yes' or 'no.' This methodology offers potentially greater efficiency compared to identifying the true label for every data instance.
As the pioneering work in complementary-label learning, \cite{ishida2017learning} assumed that the distribution of complementary labels is noise-free and can be represented by Equation~\eqref{eq:uniform}.
\begin{equation}\label{eq:uniform}
    p(\bar y = y \mid x, y) = 0,\ 
    p(x,\bar{y})=\frac{1}{K-1}\sum_{y\neq \bar{y}}p(x, y)
\end{equation}

However, assuming complementary labels to be uniformly distributed is not always realistic \cite{fwd2018}. Human annotators may exhibit biases toward specific classes and the data instance $x$. Building on this observation, studies such as \cite{fwd2018, ishiguro2022learning, cpe2023, conu2023} have expanded the loss function to accommodate biased complementary labels. These works employ a predefined, feature-independent transition matrix $T$ to represent the distribution of complementary labels conditioned on their ground-truth labels.

Additionally, there is a growing body of research focused on leveraging multiple complementary labels for supervision \cite{mcl2020, reg2021, ComCo2023, conu2023}, where each instance is assigned multiple complementary labels generated from a transition matrix without replacement. In fact, the problem of learning from multiple complementary labels can be connected to partial-label learning or negative-unlabeled learning \cite{conu2023}. Building on these concepts, \cite{clcifar2023} curated a human-labeled CIFAR~\cite{krizhevsky2009learning} dataset with complementary labels to better understand real-world CLL distributions, where the transition matrices are both biased and noisy, and each instance has three complementary labels.

There remain several open problems in CLL. First, because the transition matrix $T$ is predefined, there is no universal generation process for biased complementary labels, and a general framework is needed for fair comparison. Second, in the absence of ordinary labels, the transition matrix $T$ is often assumed to be given. If a small portion of true labels is available, the transition matrix can be estimated using the anchor point method proposed in \cite{fwd2018}. These variations can lead to inconsistent experimental outcomes. Finally, without ordinary labels, the reliability of validation using only complementary labels is uncertain.
To address these challenges, we introduce \texttt{libcll}, the first CLL toolkit, to support future CLL research and advance understanding in the weakly-supervised learning community.

\subsection{Previous methods on CLL}\label{sec:previous-methods}

\begin{figure}[htb]
\centering
    \includegraphics[width=0.80\linewidth]{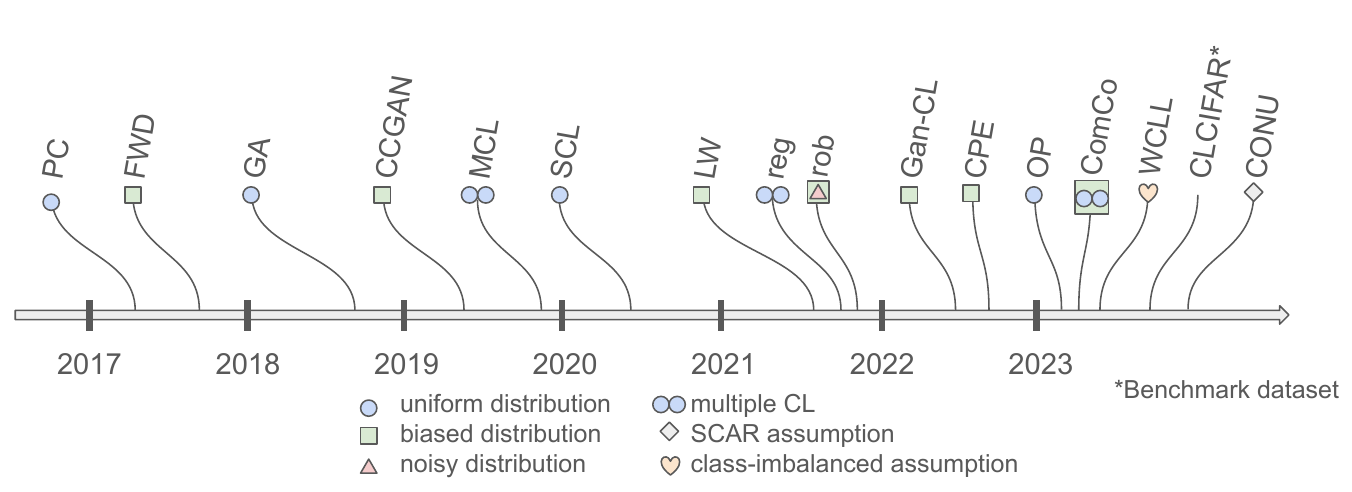}
    \caption{The development of complementary-label learning.}
    \label{fig:cll-timeline}
\end{figure}

In this section, we present a timeline of key developments in CLL, as illustrated in Figure~\ref{fig:cll-timeline}. We implement three primary categories of CLL methods in \texttt{libcll}: URE (unbiased risk estimator), CPE (complementary probability estimation), and MCL (multiple complementary label) methods. Additionally, we include several bridging works that connect CLL with other learning frameworks.

\textbf{URE Series of Works} The concept of complementary-label learning was initially proposed by \cite{ishida2017learning}, who introduced a risk estimator using Pairwise Comparison (PC) and One-vs-All (OvA) strategies for restricted loss functions. With biased complementary labels, \cite{fwd2018} employed forward correction to reconstruct classification risk using cross-entropy loss on complementary labels and the transition matrix. Furthermore, \cite{ishida2019complementarylabel} removed these restrictions, extending the unbiased risk estimator to support arbitrary loss functions and models.

\textbf{CPE Framework} \cite{cpe2023} offered a new perspective by approaching prediction as a decoding process with the transition matrix. This proposed decoding framework can unify various risk estimator methods \cite{fwd2018, scl2020, gao2021discriminative}.

\textbf{MCL Series of Works} \cite{mcl2020} introduced the concept of learning from multiple complementary labels, proposing that single-CL methods can be generalized to handle multi-CL distributions through decomposition.

Recent studies have extended the CLL framework to various fields, including Generative Adversarial Networks \cite{xu2019generativediscriminative, gancl2023}, negative-labeled learning \cite{conu2023}, robust loss functions for noisy complementary labels \cite{ishiguro2022learning}, and semi-supervised learning \cite{deng2022boosting}.
In this work, we identify the three fundamental branches of CLL: URE, CPE, and MCL, and implement these methods in \texttt{libcll}. Through extensive experiments, we validate the effectiveness of these approaches. We hope that this toolkit will inspire further advancements within the CLL community in the near future.

%% file: Tables/cll-assumptions.tex
\begin{table}[ht]
\centering
\caption{The assumptions of complementary label distribution among different CLL methods.} 
\label{tab:valid}
\vskip 0.05in
\resizebox{\columnwidth}{!}{%
\begin{tabular}{cccccccc}
\toprule
            & \multirow{2}{*}{\parbox{2cm}{\centering Include in \\ libcll}} & \multirow{2}{*}{\parbox{2cm}{\centering uniform}} & \multirow{2}{*}{\parbox{2cm}{ \centering biased}} & \multirow{2}{*}{\parbox{2cm}{\centering MCL}} & \multirow{2}{*}{\parbox{2cm}{\centering noisy}} & \multirow{2}{*}{\parbox{2cm}{\centering class-imbalanced}} & \multirow{2}{*}{\centering SCAR} \\
            & & & & & & & \\
\midrule
PC \cite{ishida2017learning} & \cmark & \cmark & & & & & \\
FWD \cite{fwd2018} & \cmark & \cmark & \cmark & & & & \\
GA \cite{ishida2019complementarylabel} & \cmark & \cmark & & & & & \\
MCL \cite{mcl2020} & \cmark & \cmark & & \cmark & & & \\
SCL \cite{scl2020} & \cmark & \cmark & \cmark & & & & \\
LW \cite{gao2021discriminative} & \cmark & \cmark & \cmark & & & & \\
rob \cite{ishiguro2022learning} & & \cmark & \cmark & & \cmark & & \\
CPE \cite{cpe2023} & \cmark & \cmark & \cmark & & & & \\
OP \cite{orderpreserving2023} & & \cmark & & & & & \\
ComCo \cite{ComCo2023} & & \cmark & \cmark & \cmark & & & \\
WCLL \cite{wei2023classimbalanced} & & \cmark & \cmark & & & \cmark & \\
CONU \cite{conu2023} & & \cmark & \cmark & \cmark & & & \cmark \\
\bottomrule 
\end{tabular}

}
\end{table}

%% file: Sections/Design.tex
The code structure of \texttt{libcll} is highly modular and seamlessly integrates with PyTorch. Each component can be added, modified, or removed individually to support diverse experimental designs. In the following paragraphs, we will outline the definitions of strategies, datasets, models, evaluation, and reproducibility.

\textbf{Strategies} in CLL algorithms are used to calculate the loss. All strategies inherit from the base class \texttt{libcll.strategies.Strategy}, which itself extends \texttt{pytorch\_lightning.LightningModule}. This means every strategy defined in \texttt{libcll} can be integrated into any other PyTorch Lightning framework. Additionally, \texttt{libcll.strategies.Strategy} already includes implementations for the validation and testing steps, as well as evaluation metrics. Users only need to create a new class that inherits from it and modify the \texttt{training\_step} to incorporate new complementary-label learning methods into the library. 
\begin{figure}[htb]
\centering
    \includegraphics[width=1.0\linewidth]{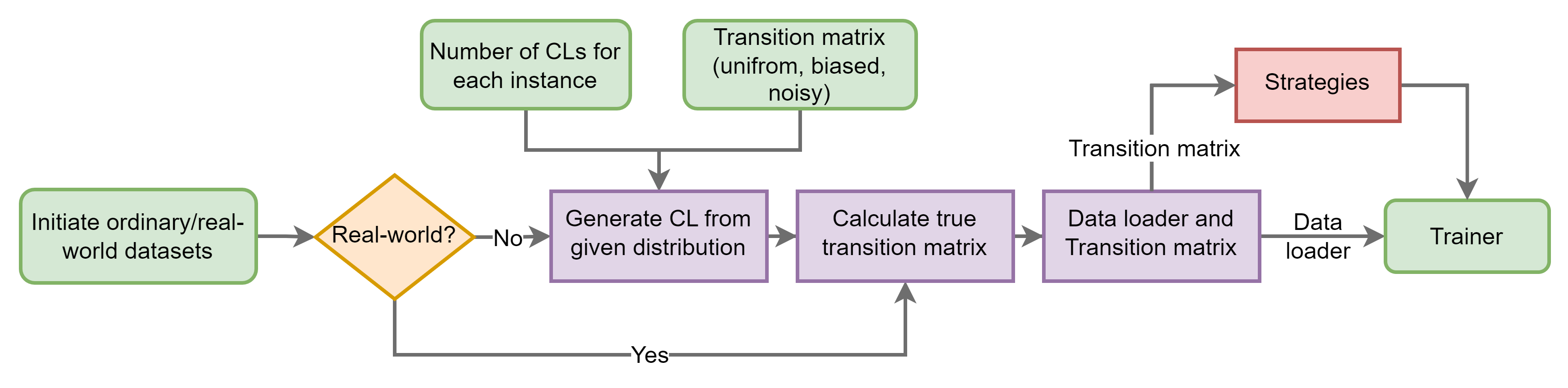}
    \caption{Training pipeline of \texttt{libcll}. The process begins by initializing the dataset, either real-world or synthetic. For real-world datasets, the pipeline directly proceeds to the calculation of the true transition matrix. If synthetic datasets are chosen, complementary labels (CLs) are generated based on user-specified parameters, such as the number of CLs per instance and their distribution (uniform, biased or noisy). The preprocessor then calculates the transition matrix based on the true and corresponding CLs for each instance. Afterward, the data preprocessor prepares the DataLoader along with the calculated transition matrix. The training process is initiated using the selected strategy module, the transition matrix, and the DataLoader.}
    \label{training_pineline}
\end{figure}

\textbf{Datasets} in \texttt{libcll} include both ordinary labels and complementary labels (CLs). During data preparation, as shown in Figure~\ref{training_pineline}, we begin by initializing an ordinary-label dataset for synthetic scenarios. Users can then generate CLs based on any distribution derived from a specified transition matrix and set the number of CLs per instance. For real-world datasets, we utilize paired data containing human-annotated CLs, ordinary labels, and images provided by~\cite{clcifar2023}, eliminating the need for CL generation. Users can select the number of human-annotated CLs per instance, up to a maximum of three.
\texttt{libcll} provides 11 synthetic datasets. Five of these (MNIST, FMNIST, KMNIST, CIFAR10, and CIFAR20) are widely used image datasets in CLL research~\cite{scl2020, ishida2019complementarylabel, gao2021discriminative, cpe2023, mcl2020, fwd2018}. The Yeast, Texture, Dermatology, and Synthetic Control datasets are from the UCI Machine Learning Repository and have been applied in multi-complementary-label learning research~\cite{mcl2020}. Additionally, MicroImageNet10 and MicroImageNet20 are recent datasets proposed by the research team in~\cite{clcifar2023}. For real-world datasets, \texttt{libcll} includes CLCIFAR10, CLCIFAR20, CLMicroImageNet10, and CLMicroImageNet20, all of which are also recent contributions from~\cite{clcifar2023}.

\textbf{Models}
\texttt{libcll} includes five models: Linear, MLP, DenseNet, ResNet18, and ResNet34. The Linear and MLP models are suitable for simpler datasets, such as MNIST, KMNIST, and FMNIST, while the DenseNet, ResNet18, and ResNet34 models are better suited for more complex datasets, such as CIFAR and (Tiny)ImageNet.

\textbf{Evaluation}
To maintain flexibility in the validation set labels and encourage future researchers to adopt more realistic setups, we provide the unbiased risk estimator (URE) and surrogate complementary estimation loss (SCEL) metrics, as proposed by~\cite{cpe2023}, for evaluation on validation sets containing only complementary labels. Additionally, we include accuracy metrics for evaluating ordinary labels.

\textbf{Reproducibility}
\texttt{libcll} saves all hyperparameters in a YAML file, along with the best model weights, and logs training loss and validation metrics at specified intervals. Additionally, \texttt{libcll} ensures consistent results when using the same hyperparameters and computational device.

%% file: Sections/Experiment.tex
\subsection{Experimental Setups}\label{sec:exp_setup} 
In the subsequent experiments in Sections~\ref{section:4.2} through~\ref{section:4.5}, we evaluate the performance of all CLL algorithms available in \texttt{libcll}. Each section utilizes distinct transition matrices and records test accuracy for each dataset and algorithm, enabling a comprehensive assessment of performance. Additionally, we calculate the average rank of test accuracy across all experiments, which is displayed in the 'Avg Rank' column of each table as the primary metric for performance comparison.
We use a one-layer MLP model (d-500-c) for the MNIST, KMNIST, FMNIST, Yeast, Texture, Control, and Dermatology datasets, and ResNet34 for the CIFAR10, CIFAR20, MicroImageNet10 (MIN10), MicroImageNet20 (MIN20), CLCIFAR10, CLCIFAR20, CLMicroImageNet10 (CLMIN10), and CLMicroImageNet20 (CLMIN20) datasets. Following the setup in \cite{clcifar2023}, we apply standard data augmentation techniques, including \texttt{RandomHorizontalFlip}, \texttt{RandomCrop}, and normalization, to each image in the CIFAR and MicroImageNet series datasets.
\input{Tables/Uniform-Simple-New}
\input{Tables/Uniform-Simple-Tabular}
\input{Tables/Biased-New}
\input{Tables/Noisy-New}

\input{Tables/Multi-Simple-New}
\input{Tables/Multi-Simple-Tabular}

Following the hyperparameters from \cite{cpe2023}, we trained each setup using the Adam optimizer with a learning rate selected from \texttt{\{1e-3, 5e-4, 1e-4, 5e-5, 1e-5\}} and a fixed batch size of 256 for 300 epochs on NVIDIA Tesla V100 32GB. We reserved 10\% of the training data as the validation set, assuming that all ground-truth labels in the validation set are known. We selected the models with the highest validation accuracy and conducted 4 trials with different random seeds for all experiments in this study.
Certain algorithms, referred to as T-aware algorithms, use a transition matrix to calculate the loss, while others, known as T-agnostic algorithms, do not utilize this information. Since T-aware algorithms benefit from additional information about the true labels, a direct comparison with T-agnostic algorithms would be inequitable. Therefore, we present and compare these algorithms separately, listing T-aware algorithms in the upper section and T-agnostic algorithms in the lower section of each table.
\subsection{Uniform distribution}
\label{section:4.2}
In this section, we establish baselines for each algorithm under the standard CLL setting, where the correct transition matrix is provided to T-agnostic methods, and complementary labels (CLs) are uniformly sampled from the complementary set. An exception is made for CLCIFAR10, CLCIFAR20, CLMIN10, and CLMIN20, whose CLs are derived from human annotations and are thus noisy. To evaluate the adaptability of current algorithms to real-world scenarios, we divide the datasets into synthetic and real-world sets.

The results are shown in Table \ref{tab:simple} and \ref{tab:simple-tab}. As observed, FWD and CPE-F demonstrate the best overall performance on both synthetic and real-world datasets, with CPE-F slightly outperforming CPE-T, consistent with findings from \cite{cpe2023}. Surprisingly, providing the transition matrix to the learner in URE-TNN and URE-TGA does not consistently yield better performance compared to URE-NN and URE-GA, particularly on non-uniform datasets such as CLCIFAR10, CLCIFAR20, CLMIN10, and CLMIN20. We suggest that the transition matrices of these datasets may be ill-conditioned, leading to instability in URE.
\subsection{Biased distribution}
\label{section:4.3}
To examine the impact of disturbances in the complementary-label distribution, we follow the procedure from \cite{gao2021discriminative} to generate two biased distributions with varying levels of deviation from a uniform distribution, as follows: For each class $y$, the complementary classes are randomly divided into three subsets, with probabilities assigned as $p_1$, $p_2$, and $p_3$ within each subset.
We consider two cases for $(p_1, p_2, p_3)$: (a) Strong: $(\frac{0.75}{3}, \frac{0.24}{3}, \frac{0.01}{3})$ to simulate a stronger deviation. (b) Weak: $(\frac{0.45}{3}, \frac{0.30}{3}, \frac{0.25}{3})$ to simulate a milder deviation.
Since these configurations are applicable only to datasets with 10 classes, we use these two distributions to generate complementary labels for MNIST, KMNIST, FMNIST, MIN10, and CIFAR10, and report the results in Table \ref{tab:biased}.

The results indicate that T-agnostic methods perform well under the Weak distribution but experience a significant accuracy drop under the Strong distribution, suggesting that these methods are sensitive to distributional deviations. In contrast, T-aware methods exhibit more stable and consistent performance. Specifically, we observe that URE-T may show a slight performance decrease when the distribution is close to uniform but achieves solid results under conditions with strong deviation from a uniform distribution.
\subsection{Noisy distribution}
\label{section:4.4}
Following the steps outlined in \cite{cpe2023}, we simulate more restrictive environments by introducing both noisy complementary labels and incorrect transition matrices to the learners. We achieve this by generating noisy datasets through the interpolation of a strong deviation matrix, $T_{\text{strong}}$, and a uniform noise transition matrix, $\frac{1}{K}1_K$.
The resulting complementary labels follow the distribution $(1 - \lambda)T_{\text{strong}} + \lambda \frac{1}{K}1_K$, while only $T_{\text{strong}}$ is provided to the learners, where $\lambda$ controls the weight of the noise.

The results, presented in Table \ref{tab:noisy}, indicate a performance drop across all methods, particularly as the noise factor $\lambda$ increases. This demonstrates that, while T-aware methods can manage some degree of deviation and noise in the transition matrix, they become increasingly vulnerable as the gap widens between the provided transition matrix and the actual distribution with higher noise levels. This highlights why CPE-T outperforms other T-aware methods in noisy settings, as it includes a trainable transition layer that mitigates this gap.
\subsection{Multi-label distribution}
\label{section:4.5}
To demonstrate the versatility of CLL, we assign three complementary labels to each instance, sampled from a uniform distribution without repetition. For real-world datasets, we use three human-annotated complementary labels per sample. After generating multiple complementary labels, one-hot encoding is applied for MCL series loss calculations. For other loss functions, instances are decomposed into multiple examples, each containing a single complementary label, with a shuffled training set.

The results in Table~\ref{tab:multi-simple} reveal that the performance of MCL improves significantly, achieving the highest accuracy in synthetic multiple-CL scenarios. However, as shown in Table~\ref{tab:multi-simple-tabular}, the noisy distribution in real-world datasets continues to impair the performance of T-agnostic algorithms. Additionally, as the use of multiple complementary labels increases the noise level in real-world datasets, the performance of both T-agnostic and T-aware algorithms does not improve—or may even decline—compared to single-CL scenarios on the same datasets. This suggests that simply increasing the number of complementary labels is insufficient to effectively address the challenges posed by noisy distributions.

%% file: Tables/Uniform-Simple-New.tex
\begin{table}[htb]
\centering
\caption{Baseline testing accuracies of various strategies with a uniform distribution on synthetic datasets. Results are grouped into two categories: T-agnostic algorithms (upper section) and T-aware algorithms (lower section). The \textit{best performance} for each dataset is highlighted in bold, while the \textit{second-best} result is underlined. The 'Avg Rank' column provides an overall ranking of the algorithms across all datasets, where a lower value indicates better general performance.}
\label{tab:simple}
\scalebox{0.75}{
\begin{tabular}{lcccccccccccccccccccccccccc}
\toprule
& {MNIST} & & {KMNIST} & & {FMNIST} & & {CIFAR10} & & {CIFAR20} & & {MIN10} & & {MIN20} & & Avg Rank \\
\midrule
SCL-NL & 94.20\scriptsize{$\pm$0.10} & & \underline{71.88}\scriptsize{$\pm$0.45} & & \textbf{84.59}\scriptsize{$\pm$0.41} & & \underline{65.45}\scriptsize{$\pm$0.75} & & 21.43\scriptsize{$\pm$1.00} & & 37.59\scriptsize{$\pm$2.96} & & \underline{11.75}\scriptsize{$\pm$2.36} & & \underline{2.57}\\
SCL-EXP & 94.29\scriptsize{$\pm$0.29} & & 67.82\scriptsize{$\pm$0.52} & & 84.30\scriptsize{$\pm$0.32} & & 61.16\scriptsize{$\pm$1.18} & & 21.33\scriptsize{$\pm$0.20} & & \underline{39.14}\scriptsize{$\pm$3.49} & & 10.83\scriptsize{$\pm$2.32} & & 3.86\\
URE-NN & 92.19\scriptsize{$\pm$0.04} & & 69.27\scriptsize{$\pm$1.24} & & 82.73\scriptsize{$\pm$0.27} & & 46.83\scriptsize{$\pm$1.41} & & 15.48\scriptsize{$\pm$0.67} & & 29.46\scriptsize{$\pm$1.85} & & 10.98\scriptsize{$\pm$0.83} & & 5.71\\
URE-GA & \textbf{94.41}\scriptsize{$\pm$0.16} & & \textbf{74.32}\scriptsize{$\pm$0.45} & & 83.51\scriptsize{$\pm$0.22} & & 58.00\scriptsize{$\pm$1.49} & & 15.18\scriptsize{$\pm$0.47} & & 33.70\scriptsize{$\pm$1.34} & & 8.62\scriptsize{$\pm$1.68} & & 4.71\\
DM & 92.41\scriptsize{$\pm$0.17} & & 68.72\scriptsize{$\pm$0.92} & & 82.67\scriptsize{$\pm$0.53} & & 54.04\scriptsize{$\pm$1.26} & & 18.45\scriptsize{$\pm$0.97} & & 36.49\scriptsize{$\pm$4.66} & & 8.85\scriptsize{$\pm$1.33} & & 5.86\\
MCL-MAE & 91.84\scriptsize{$\pm$3.95} & & 64.33\scriptsize{$\pm$3.61} & & 82.63\scriptsize{$\pm$0.31} & & 38.07\scriptsize{$\pm$9.22} & & 11.65\scriptsize{$\pm$1.77} & & 26.66\scriptsize{$\pm$3.60} & & 9.07\scriptsize{$\pm$1.64} & & 7.57\\
MCL-EXP & 94.24\scriptsize{$\pm$0.27} & & 68.25\scriptsize{$\pm$0.56} & & 84.03\scriptsize{$\pm$0.58} & & 61.19\scriptsize{$\pm$1.05} & & \underline{21.45}\scriptsize{$\pm$0.93} & & 37.44\scriptsize{$\pm$2.65} & & 9.05\scriptsize{$\pm$1.09} & & 4.14\\
MCL-LOG & \underline{94.31}\scriptsize{$\pm$0.16} & & 70.52\scriptsize{$\pm$2.42} & & \underline{84.50}\scriptsize{$\pm$0.32} & & \textbf{66.25}\scriptsize{$\pm$0.66} & & \textbf{21.59}\scriptsize{$\pm$1.11} & & \textbf{40.04}\scriptsize{$\pm$2.71} & & \textbf{11.89}\scriptsize{$\pm$1.30} & & \bf{1.57}\\
\midrule
FWD & \underline{93.78}\scriptsize{$\pm$0.35} & & \underline{70.16}\scriptsize{$\pm$1.56} & & \textbf{84.16}\scriptsize{$\pm$0.25} & & \textbf{64.99}\scriptsize{$\pm$0.64} & & 20.50\scriptsize{$\pm$0.75} & & \underline{36.72}\scriptsize{$\pm$4.30} & & \underline{13.35}\scriptsize{$\pm$2.40} & & \bf{1.86}\\
URE-TNN & 91.85\scriptsize{$\pm$0.23} & & 69.16\scriptsize{$\pm$1.14} & & 82.10\scriptsize{$\pm$0.30} & & 47.25\scriptsize{$\pm$2.35} & & 15.29\scriptsize{$\pm$1.11} & & 30.52\scriptsize{$\pm$2.32} & & 10.33\scriptsize{$\pm$1.33} & & 5.00\\
URE-TGA & \textbf{93.79}\scriptsize{$\pm$0.14} & & \textbf{75.53}\scriptsize{$\pm$1.10} & & 83.76\scriptsize{$\pm$0.39} & & 57.37\scriptsize{$\pm$1.20} & & 13.61\scriptsize{$\pm$4.51} & & 34.32\scriptsize{$\pm$2.53} & & 6.33\scriptsize{$\pm$2.46} & & 3.43\\
CPE-I & 91.30\scriptsize{$\pm$0.25} & & 64.82\scriptsize{$\pm$0.64} & & 81.25\scriptsize{$\pm$0.77} & & 54.44\scriptsize{$\pm$1.95} & & 12.01\scriptsize{$\pm$0.93} & & 28.98\scriptsize{$\pm$1.16} & & 10.57\scriptsize{$\pm$1.36} & & 5.57\\
CPE-F & 93.77\scriptsize{$\pm$0.35} & & 70.15\scriptsize{$\pm$1.57} & & \underline{83.95}\scriptsize{$\pm$0.34} & & \underline{64.97}\scriptsize{$\pm$0.61} & & \textbf{20.73}\scriptsize{$\pm$1.17} & & 36.46\scriptsize{$\pm$4.06} & & \textbf{14.98}\scriptsize{$\pm$1.53} & & \underline{2.14}\\
CPE-T & 93.40\scriptsize{$\pm$0.27} & & 69.75\scriptsize{$\pm$1.49} & & 83.66\scriptsize{$\pm$0.14} & & 57.94\scriptsize{$\pm$0.97} & & \underline{20.58}\scriptsize{$\pm$0.78} & & \textbf{37.16}\scriptsize{$\pm$2.82} & & 13.14\scriptsize{$\pm$2.10} & & 3.00\\
\bottomrule
\end{tabular}%
}
\end{table}

%% file: Tables/Uniform-Simple-Tabular.tex
\begin{table}[ht]
\centering
\caption{Baseline testing accuracies of strategies with uniform distribution on Synthetic datasets and Real-World datasets.}
\label{tab:simple-tab}
\resizebox{1\linewidth}{!}{
\begin{tabular}{lcccccccccccccccccccccccccc}
\toprule
& \multicolumn{5}{c}{Synthetic Tabular Datasets} & \multicolumn{5}{c}{Real-World Datasets} \\
\cmidrule(lr){2-6} \cmidrule(lr){7-11}
 & {Yeast} & {Texture} & {Dermatology} & {Control} & Avg Rank & {CLCIFAR10} & {CLCIFAR20} & {CLMIN10} & {CLMIN20} & {Avg Rank} \\
\midrule
SCL-NL & 49.33\scriptsize{$\pm$5.46} & 97.22\scriptsize{$\pm$0.83} & 72.97\scriptsize{$\pm$6.04} & \underline{77.50}\scriptsize{$\pm$10.57} & \underline{3.25} &  36.21\scriptsize{$\pm$2.06} & \underline{7.92}\scriptsize{$\pm$0.23} & \underline{16.59}\scriptsize{$\pm$3.46} & \textbf{6.82}\scriptsize{$\pm$0.53} & \bf{2.25}\\
SCL-EXP & 47.15\scriptsize{$\pm$6.44} & \textbf{97.40}\scriptsize{$\pm$0.79} & 70.27\scriptsize{$\pm$6.62} & 77.08\scriptsize{$\pm$11.39} & 4.38 &  35.95\scriptsize{$\pm$1.97} & 7.79\scriptsize{$\pm$0.06} & 12.60\scriptsize{$\pm$0.58} & 5.81\scriptsize{$\pm$0.86} & 4.75\\
URE-NN & 48.66\scriptsize{$\pm$2.15} & 86.95\scriptsize{$\pm$3.77} & \textbf{85.14}\scriptsize{$\pm$2.34} & 51.25\scriptsize{$\pm$6.17} & 4.25 &  34.66\scriptsize{$\pm$1.64} & \textbf{11.77}\scriptsize{$\pm$0.72} & \textbf{19.51}\scriptsize{$\pm$3.21} & 4.93\scriptsize{$\pm$0.42} & 4.00\\
URE-GA & 47.82\scriptsize{$\pm$3.33} & 79.59\scriptsize{$\pm$5.87} & \underline{79.05}\scriptsize{$\pm$6.16} & 45.42\scriptsize{$\pm$5.45} & 5.25 &  \underline{37.32}\scriptsize{$\pm$0.61} & 7.09\scriptsize{$\pm$0.77} & 14.34\scriptsize{$\pm$2.48} & \underline{5.99}\scriptsize{$\pm$0.92} & \underline{3.50}\\
DM & 34.73\scriptsize{$\pm$7.45} & 95.34\scriptsize{$\pm$1.32} & 70.27\scriptsize{$\pm$6.89} & \textbf{78.75}\scriptsize{$\pm$8.53} & 4.88 &  36.78\scriptsize{$\pm$0.81} & 7.66\scriptsize{$\pm$0.10} & 14.03\scriptsize{$\pm$1.69} & 5.48\scriptsize{$\pm$0.74} & 4.50\\
MCL-MAE & 33.72\scriptsize{$\pm$5.66} & 60.51\scriptsize{$\pm$9.40} & 66.22\scriptsize{$\pm$11.55} & 43.33\scriptsize{$\pm$7.73} & 8.00 &  20.01\scriptsize{$\pm$1.41} & 6.55\scriptsize{$\pm$0.76} & 11.94\scriptsize{$\pm$1.00} & 5.78\scriptsize{$\pm$0.54} & 7.25\\
MCL-EXP & \underline{51.01}\scriptsize{$\pm$6.49} & \underline{97.28}\scriptsize{$\pm$1.00} & 73.65\scriptsize{$\pm$5.85} & 77.50\scriptsize{$\pm$10.96} & \bf{2.62} &  34.54\scriptsize{$\pm$1.09} & 7.82\scriptsize{$\pm$0.23} & 13.75\scriptsize{$\pm$2.39} & 5.43\scriptsize{$\pm$1.03} & 5.75\\
MCL-LOG & \textbf{55.37}\scriptsize{$\pm$6.30} & 96.93\scriptsize{$\pm$1.05} & 73.65\scriptsize{$\pm$7.73} & 76.67\scriptsize{$\pm$7.55} & 3.38 &  \textbf{37.35}\scriptsize{$\pm$1.17} & 7.49\scriptsize{$\pm$0.39} & 13.93\scriptsize{$\pm$2.75} & 5.79\scriptsize{$\pm$0.47} & 4.00\\
\midrule
FWD & \textbf{58.72}\scriptsize{$\pm$3.17} & \textbf{97.12}\scriptsize{$\pm$0.75} & 79.05\scriptsize{$\pm$9.63} & \textbf{90.83}\scriptsize{$\pm$6.72} & \bf{1.75} &  \textbf{39.25}\scriptsize{$\pm$1.55} & \textbf{19.82}\scriptsize{$\pm$0.38} & \underline{29.63}\scriptsize{$\pm$2.92} & \textbf{10.58}\scriptsize{$\pm$1.14} & \bf{1.25}\\
URE-TNN & 37.75\scriptsize{$\pm$14.53} & 88.01\scriptsize{$\pm$3.08} & \underline{81.08}\scriptsize{$\pm$3.82} & 62.92\scriptsize{$\pm$5.45} & 4.00 &  31.21\scriptsize{$\pm$1.11} & 9.47\scriptsize{$\pm$2.67} & 17.87\scriptsize{$\pm$4.33} & 6.30\scriptsize{$\pm$0.96} & 5.50\\
URE-TGA & 35.57\scriptsize{$\pm$15.60} & 81.42\scriptsize{$\pm$2.74} & 78.38\scriptsize{$\pm$5.06} & 58.75\scriptsize{$\pm$6.28} & 6.00 &  33.68\scriptsize{$\pm$1.06} & 5.17\scriptsize{$\pm$0.25} & 21.35\scriptsize{$\pm$4.38} & 5.88\scriptsize{$\pm$1.21} & 5.25\\
CPE-I & 52.68\scriptsize{$\pm$1.93} & 91.39\scriptsize{$\pm$0.86} & \textbf{81.76}\scriptsize{$\pm$5.85} & 59.17\scriptsize{$\pm$13.15} & 3.50 &  33.94\scriptsize{$\pm$1.40} & 16.79\scriptsize{$\pm$0.48} & 20.18\scriptsize{$\pm$4.25} & 8.61\scriptsize{$\pm$0.43} & 4.25\\
CPE-F & 56.04\scriptsize{$\pm$3.84} & \underline{96.67}\scriptsize{$\pm$0.89} & 79.73\scriptsize{$\pm$2.34} & \underline{90.00}\scriptsize{$\pm$6.24} & \underline{2.50} &  \underline{38.94}\scriptsize{$\pm$1.39} & \underline{19.48}\scriptsize{$\pm$0.35} & \textbf{29.82}\scriptsize{$\pm$3.13} & \underline{10.16}\scriptsize{$\pm$0.84} & \underline{1.75}\\
CPE-T & \underline{57.89}\scriptsize{$\pm$2.62} & 95.60\scriptsize{$\pm$0.53} & 79.05\scriptsize{$\pm$1.17} & 87.92\scriptsize{$\pm$5.94} & 3.25 &  38.80\scriptsize{$\pm$1.16} & 19.33\scriptsize{$\pm$0.75} & 28.85\scriptsize{$\pm$0.55} & 9.36\scriptsize{$\pm$1.32} & 3.00\\
\bottomrule
\end{tabular}%
}
\end{table}

%% file: Tables/Biased-New.tex
\begin{table}[ht]
\centering
\caption{Comparison of testing accuracies under Weak and Strong levels of deviation in transition matrices on synthetic datasets. The definition of Weak and Strong deviation are described in section \ref{section:4.3}}.
\label{tab:biased}
\resizebox{1\linewidth}{!}{%
\begin{tabular}{lcccccccccccccccccccc}
\toprule
& \multicolumn{2}{c}{MNIST} & \multicolumn{2}{c}{KMNIST} & \multicolumn{2}{c}{FMNIST} & \multicolumn{2}{c}{CIFAR10} & \multicolumn{2}{c}{MIN10} & \multicolumn{2}{c}{Avg Rank} \\
\cmidrule(lr){2-3} \cmidrule(lr){4-5} \cmidrule(lr){6-7} \cmidrule(lr){8-9} \cmidrule(lr){10-11} \cmidrule(lr){12-13}
& Weak & Strong & Weak & Strong & Weak & Strong & Weak & Strong & Weak & Strong & Weak & Strong \\
\midrule
SCL-NL & 93.94\scriptsize{$\pm$0.34} & 33.29\scriptsize{$\pm$7.21} & 70.44\scriptsize{$\pm$2.20} & 26.40\scriptsize{$\pm$3.89} & \textbf{82.92}\scriptsize{$\pm$0.31} & 27.93\scriptsize{$\pm$3.75} & 58.63\scriptsize{$\pm$2.78} & 24.01\scriptsize{$\pm$4.27} & \underline{37.18}\scriptsize{$\pm$0.68} & 15.64\scriptsize{$\pm$3.75} & 3.00 & \underline{3.20}\\
SCL-EXP & 94.05\scriptsize{$\pm$0.30} & 27.64\scriptsize{$\pm$5.29} & 68.88\scriptsize{$\pm$2.00} & 22.49\scriptsize{$\pm$2.56} & 79.65\scriptsize{$\pm$3.86} & 24.43\scriptsize{$\pm$3.67} & 58.71\scriptsize{$\pm$3.24} & 19.03\scriptsize{$\pm$2.52} & 35.98\scriptsize{$\pm$2.76} & 14.12\scriptsize{$\pm$4.50} & 4.60 & 5.60\\
URE-NN & 92.17\scriptsize{$\pm$0.38} & \textbf{48.66}\scriptsize{$\pm$7.65} & 69.98\scriptsize{$\pm$0.42} & \textbf{33.62}\scriptsize{$\pm$3.98} & 82.08\scriptsize{$\pm$0.57} & \textbf{41.57}\scriptsize{$\pm$4.05} & 50.09\scriptsize{$\pm$2.20} & \underline{31.31}\scriptsize{$\pm$2.63} & 29.04\scriptsize{$\pm$2.11} & \textbf{23.04}\scriptsize{$\pm$3.82} & 5.60 & \textbf{1.20}\\
URE-GA & \textbf{94.49}\scriptsize{$\pm$0.23} & 33.26\scriptsize{$\pm$14.30} & \textbf{75.46}\scriptsize{$\pm$0.64} & 23.81\scriptsize{$\pm$4.71} & 82.28\scriptsize{$\pm$0.51} & 20.85\scriptsize{$\pm$3.33} & \underline{59.66}\scriptsize{$\pm$0.89} & 18.45\scriptsize{$\pm$5.75} & 33.44\scriptsize{$\pm$2.22} & 16.63\scriptsize{$\pm$4.24} & \underline{2.40} & 5.00\\
DM & 91.24\scriptsize{$\pm$0.27} & 31.05\scriptsize{$\pm$4.02} & 67.33\scriptsize{$\pm$1.66} & 22.18\scriptsize{$\pm$2.51} & 80.55\scriptsize{$\pm$0.37} & 26.36\scriptsize{$\pm$2.21} & 55.40\scriptsize{$\pm$2.03} & \textbf{32.92}\scriptsize{$\pm$4.55} & 32.26\scriptsize{$\pm$1.23} & \underline{21.04}\scriptsize{$\pm$3.44} & 6.20 & 4.00\\
MCL-MAE & 67.61\scriptsize{$\pm$7.37} & 20.37\scriptsize{$\pm$6.82} & 48.64\scriptsize{$\pm$6.40} & 18.08\scriptsize{$\pm$2.23} & 67.01\scriptsize{$\pm$9.04} & 19.91\scriptsize{$\pm$2.34} & 21.08\scriptsize{$\pm$4.87} & 12.89\scriptsize{$\pm$2.70} & 18.63\scriptsize{$\pm$5.63} & 11.85\scriptsize{$\pm$0.97} & 8.00 & 8.00\\
MCL-EXP & 94.06\scriptsize{$\pm$0.27} & 26.76\scriptsize{$\pm$5.43} & 68.60\scriptsize{$\pm$1.35} & 22.65\scriptsize{$\pm$2.59} & 79.82\scriptsize{$\pm$3.87} & 27.03\scriptsize{$\pm$2.91} & 56.76\scriptsize{$\pm$3.66} & 18.80\scriptsize{$\pm$5.52} & 36.61\scriptsize{$\pm$0.92} & 13.82\scriptsize{$\pm$4.15} & 4.60 & 5.60\\
MCL-LOG & \underline{94.17}\scriptsize{$\pm$0.23} & \underline{33.65}\scriptsize{$\pm$7.26} & \underline{71.12}\scriptsize{$\pm$2.27} & \underline{26.88}\scriptsize{$\pm$4.17} & \underline{82.61}\scriptsize{$\pm$0.37} & \underline{28.07}\scriptsize{$\pm$3.98} & \textbf{63.41}\scriptsize{$\pm$1.70} & 22.92\scriptsize{$\pm$3.53} & \textbf{37.50}\scriptsize{$\pm$1.82} & 13.31\scriptsize{$\pm$4.22} & \textbf{1.60} & 3.40\\
\midrule
FWD & \textbf{93.69}\scriptsize{$\pm$0.07} & \textbf{96.23}\scriptsize{$\pm$0.13} & \underline{70.82}\scriptsize{$\pm$2.10} & \textbf{82.34}\scriptsize{$\pm$0.49} & \textbf{83.77}\scriptsize{$\pm$0.52} & \textbf{86.66}\scriptsize{$\pm$0.25} & \textbf{66.38}\scriptsize{$\pm$2.16} & \textbf{80.04}\scriptsize{$\pm$1.88} & \textbf{38.35}\scriptsize{$\pm$2.86} & \underline{53.20}\scriptsize{$\pm$2.62} & \textbf{1.20} & \textbf{1.20}\\
URE-TNN & 90.28\scriptsize{$\pm$0.79} & 90.85\scriptsize{$\pm$0.63} & 65.96\scriptsize{$\pm$0.85} & 65.71\scriptsize{$\pm$4.11} & 81.04\scriptsize{$\pm$0.85} & 82.15\scriptsize{$\pm$0.93} & 43.82\scriptsize{$\pm$2.04} & 45.30\scriptsize{$\pm$8.13} & 24.13\scriptsize{$\pm$2.69} & 25.29\scriptsize{$\pm$4.44} & 5.40 & 5.60\\
URE-TGA & 92.49\scriptsize{$\pm$0.50} & 90.79\scriptsize{$\pm$3.25} & \textbf{71.52}\scriptsize{$\pm$1.09} & 70.52\scriptsize{$\pm$5.30} & 82.52\scriptsize{$\pm$0.61} & 64.98\scriptsize{$\pm$31.88} & 56.62\scriptsize{$\pm$1.05} & 49.13\scriptsize{$\pm$23.02} & 31.96\scriptsize{$\pm$1.78} & 35.01\scriptsize{$\pm$9.76} & 3.60 & 5.40\\
CPE-I & 89.27\scriptsize{$\pm$0.26} & 93.15\scriptsize{$\pm$0.50} & 63.84\scriptsize{$\pm$1.20} & 74.39\scriptsize{$\pm$1.43} & 80.40\scriptsize{$\pm$0.30} & 83.91\scriptsize{$\pm$1.17} & 60.20\scriptsize{$\pm$1.35} & 77.11\scriptsize{$\pm$1.67} & 28.24\scriptsize{$\pm$1.42} & 43.72\scriptsize{$\pm$2.00} & 5.40 & 3.80\\
CPE-F & \underline{93.63}\scriptsize{$\pm$0.06} & \underline{96.08}\scriptsize{$\pm$0.23} & 70.78\scriptsize{$\pm$2.04} & \underline{82.13}\scriptsize{$\pm$0.60} & \underline{83.58}\scriptsize{$\pm$0.48} & \underline{86.44}\scriptsize{$\pm$0.46} & \underline{66.24}\scriptsize{$\pm$2.04} & \underline{79.59}\scriptsize{$\pm$1.87} & 36.74\scriptsize{$\pm$2.38} & 52.85\scriptsize{$\pm$2.56} & \underline{2.40} & \underline{2.20}\\
CPE-T & 93.59\scriptsize{$\pm$0.10} & 96.07\scriptsize{$\pm$0.24} & 70.16\scriptsize{$\pm$2.17} & 82.00\scriptsize{$\pm$0.61} & 83.29\scriptsize{$\pm$0.47} & 85.70\scriptsize{$\pm$0.76} & 64.26\scriptsize{$\pm$1.92} & 76.68\scriptsize{$\pm$3.27} & \underline{36.82}\scriptsize{$\pm$1.09} & \textbf{54.65}\scriptsize{$\pm$2.04} & 3.00 & 2.80\\
\bottomrule
\end{tabular}%
}
\end{table}

%% file: Tables/Noisy-New.tex
\begin{table}[ht]
\centering
\caption{Comparison of testing accuracies with different levels of noise in transition matrices. $\lambda$ stands for the weights between uniform noise transition matrix $\frac{1}{K}1_K$ and Strong deviation transition matrix. Datasets with higher $\lambda$ are noisier.}
\label{tab:noisy}
\resizebox{1\linewidth}{!}{
\begin{tabular}{lcccccccccccccccccccccccc}
\toprule
& \multicolumn{2}{c}{MNIST} & \multicolumn{2}{c}{KMNIST} & \multicolumn{2}{c}{FMNIST} & \multicolumn{2}{c}{CIFAR10} & \multicolumn{2}{c}{MIN10} & \multicolumn{2}{c}{Avg Rank} \\
\cmidrule(lr){2-3} \cmidrule(lr){4-5} \cmidrule(lr){6-7} \cmidrule(lr){8-9} \cmidrule(lr){10-11} \cmidrule(lr){12-13}
& $\lambda = 0.1$ & $\lambda = 0.5$ & $\lambda = 0.1$ & $\lambda = 0.5$ & $\lambda = 0.1$ & $\lambda = 0.5$ & $\lambda = 0.1$ & $\lambda = 0.5$ & $\lambda = 0.1$ & $\lambda = 0.5$ & $\lambda = 0.1$ & $\lambda = 0.5$ \\
\midrule
SCL-NL & 24.97\scriptsize{$\pm$5.78} & 27.12\scriptsize{$\pm$5.47} & 19.27\scriptsize{$\pm$3.75} & 18.90\scriptsize{$\pm$2.55} & 22.72\scriptsize{$\pm$2.74} & \underline{25.61}\scriptsize{$\pm$3.09} & 20.70\scriptsize{$\pm$2.82} & 20.29\scriptsize{$\pm$0.27} & 18.94\scriptsize{$\pm$4.80} & 15.75\scriptsize{$\pm$3.13} & 4.80 & 3.80\\
SCL-EXP & 18.24\scriptsize{$\pm$8.41} & 24.20\scriptsize{$\pm$5.27} & 20.36\scriptsize{$\pm$2.88} & 19.12\scriptsize{$\pm$2.60} & 21.04\scriptsize{$\pm$1.91} & 23.97\scriptsize{$\pm$2.82} & 19.78\scriptsize{$\pm$3.39} & 19.89\scriptsize{$\pm$1.40} & 13.80\scriptsize{$\pm$1.80} & 14.25\scriptsize{$\pm$1.22} & 6.20 & 5.60\\
URE-NN & \textbf{40.09}\scriptsize{$\pm$3.61} & 26.81\scriptsize{$\pm$2.94} & \textbf{31.80}\scriptsize{$\pm$5.54} & \textbf{23.70}\scriptsize{$\pm$3.66} & \textbf{34.61}\scriptsize{$\pm$6.09} & \textbf{30.94}\scriptsize{$\pm$2.70} & \textbf{32.30}\scriptsize{$\pm$4.64} & \textbf{22.42}\scriptsize{$\pm$1.99} & \textbf{22.55}\scriptsize{$\pm$3.71} & \textbf{20.50}\scriptsize{$\pm$1.58} & \textbf{1.00} & \textbf{1.60}\\
URE-GA & \underline{34.89}\scriptsize{$\pm$9.62} & \underline{27.15}\scriptsize{$\pm$4.16} & \underline{23.88}\scriptsize{$\pm$3.84} & \underline{20.49}\scriptsize{$\pm$3.94} & 26.05\scriptsize{$\pm$3.47} & 22.03\scriptsize{$\pm$1.38} & 21.97\scriptsize{$\pm$4.14} & 20.39\scriptsize{$\pm$0.72} & 18.17\scriptsize{$\pm$4.59} & \underline{17.59}\scriptsize{$\pm$3.29} & 2.80 & \underline{3.20}\\
DM & 27.71\scriptsize{$\pm$3.70} & \textbf{28.60}\scriptsize{$\pm$2.84} & 22.05\scriptsize{$\pm$1.57} & 19.46\scriptsize{$\pm$1.01} & \underline{26.14}\scriptsize{$\pm$3.70} & 25.31\scriptsize{$\pm$1.04} & \underline{22.39}\scriptsize{$\pm$0.47} & 18.88\scriptsize{$\pm$0.95} & \underline{20.30}\scriptsize{$\pm$3.69} & 15.16\scriptsize{$\pm$1.59} & \underline{2.40} & 3.80\\
MCL-MAE & 19.45\scriptsize{$\pm$5.32} & 21.07\scriptsize{$\pm$2.35} & 17.08\scriptsize{$\pm$2.69} & 17.70\scriptsize{$\pm$2.93} & 18.59\scriptsize{$\pm$4.96} & 18.09\scriptsize{$\pm$5.79} & 12.96\scriptsize{$\pm$3.32} & 12.51\scriptsize{$\pm$2.91} & 11.22\scriptsize{$\pm$0.55} & 11.79\scriptsize{$\pm$1.18} & 7.40 & 8.00\\
MCL-EXP & 18.23\scriptsize{$\pm$8.42} & 24.12\scriptsize{$\pm$5.57} & 20.24\scriptsize{$\pm$2.93} & 19.00\scriptsize{$\pm$2.61} & 17.70\scriptsize{$\pm$2.74} & 23.20\scriptsize{$\pm$2.76} & 20.55\scriptsize{$\pm$2.84} & 18.28\scriptsize{$\pm$1.81} & 16.58\scriptsize{$\pm$4.92} & 14.63\scriptsize{$\pm$2.53} & 6.60 & 6.40\\
MCL-LOG & 22.44\scriptsize{$\pm$7.10} & 26.28\scriptsize{$\pm$5.96} & 19.99\scriptsize{$\pm$3.05} & 19.77\scriptsize{$\pm$2.75} & 24.31\scriptsize{$\pm$4.30} & 24.43\scriptsize{$\pm$2.48} & 20.93\scriptsize{$\pm$3.33} & \underline{20.39}\scriptsize{$\pm$1.69} & 17.87\scriptsize{$\pm$3.42} & 15.69\scriptsize{$\pm$2.00} & 4.80 & 3.60\\
\midrule
FWD & \textbf{95.17}\scriptsize{$\pm$0.21} & 84.27\scriptsize{$\pm$3.98} & \textbf{79.56}\scriptsize{$\pm$0.43} & 60.83\scriptsize{$\pm$3.83} & \textbf{85.32}\scriptsize{$\pm$0.11} & 74.84\scriptsize{$\pm$2.03} & \textbf{75.21}\scriptsize{$\pm$1.02} & 52.73\scriptsize{$\pm$2.96} & \textbf{50.10}\scriptsize{$\pm$3.54} & \textbf{30.76}\scriptsize{$\pm$1.61} & \textbf{1.00} & 3.40\\
URE-TNN & 89.62\scriptsize{$\pm$1.01} & 78.13\scriptsize{$\pm$3.18} & 64.66\scriptsize{$\pm$2.88} & 47.14\scriptsize{$\pm$5.24} & 80.52\scriptsize{$\pm$1.44} & 74.54\scriptsize{$\pm$1.02} & 46.44\scriptsize{$\pm$2.28} & 27.91\scriptsize{$\pm$3.09} & 27.21\scriptsize{$\pm$0.85} & 20.83\scriptsize{$\pm$2.29} & 6.00 & 6.00\\
URE-TGA & 91.94\scriptsize{$\pm$0.58} & 83.45\scriptsize{$\pm$2.36} & 69.39\scriptsize{$\pm$2.46} & 49.87\scriptsize{$\pm$5.56} & 82.88\scriptsize{$\pm$0.77} & 76.36\scriptsize{$\pm$1.41} & 49.17\scriptsize{$\pm$14.47} & 33.08\scriptsize{$\pm$3.18} & 33.13\scriptsize{$\pm$2.64} & 24.61\scriptsize{$\pm$1.19} & 5.00 & 4.80\\
CPE-I & 92.95\scriptsize{$\pm$0.51} & \underline{87.92}\scriptsize{$\pm$0.61} & 72.65\scriptsize{$\pm$1.57} & 60.77\scriptsize{$\pm$2.20} & 83.22\scriptsize{$\pm$1.20} & \textbf{78.78}\scriptsize{$\pm$1.05} & 73.48\scriptsize{$\pm$2.06} & 54.94\scriptsize{$\pm$1.20} & 41.46\scriptsize{$\pm$0.97} & 27.58\scriptsize{$\pm$2.44} & 4.00 & 2.80\\
CPE-F & \underline{95.16}\scriptsize{$\pm$0.39} & 87.66\scriptsize{$\pm$2.36} & \underline{79.41}\scriptsize{$\pm$0.79} & \underline{63.15}\scriptsize{$\pm$3.38} & \underline{84.44}\scriptsize{$\pm$0.73} & 77.84\scriptsize{$\pm$2.02} & 74.21\scriptsize{$\pm$1.04} & \underline{55.41}\scriptsize{$\pm$2.34} & \underline{49.56}\scriptsize{$\pm$3.28} & 28.15\scriptsize{$\pm$2.15} & \underline{2.20} & \underline{2.60}\\
CPE-T & 94.93\scriptsize{$\pm$0.43} & \textbf{87.93}\scriptsize{$\pm$1.77} & 79.24\scriptsize{$\pm$0.99} & \textbf{63.19}\scriptsize{$\pm$3.55} & 84.12\scriptsize{$\pm$1.09} & \underline{78.55}\scriptsize{$\pm$2.04} & \underline{74.89}\scriptsize{$\pm$1.10} & \textbf{55.94}\scriptsize{$\pm$1.54} & 42.93\scriptsize{$\pm$1.22} & \underline{28.42}\scriptsize{$\pm$1.29} & 2.80 & \textbf{1.40}\\
\bottomrule
\end{tabular}
}
\end{table}

%% file: Tables/Multi-Simple-New.tex
\begin{table}[t]
\centering
\caption{Comparison of test accuracies where instances have 3 CLs on Synthetic datasets.}
\label{tab:multi-simple}
\fontsize{9.5pt}{10.25pt}\selectfont
\scalebox{0.75}{
\begin{tabular}{lccccccccccccccccccccccccccc}
\toprule
& {MNIST} & &  {KMNIST} & &  {FMNIST} & &  {CIFAR10} & &  {CIFAR20} & &  {MIN10} & &  {MIN20} & &  {Avg Rank} \\
\midrule
SCL-NL &  \textbf{96.83}\scriptsize{$\pm$0.08} & &  \textbf{81.97}\scriptsize{$\pm$2.19} & &  \textbf{86.74}\scriptsize{$\pm$0.26} & &  \textbf{82.76}\scriptsize{$\pm$0.15} & &  \textbf{34.64}\scriptsize{$\pm$0.37} & &  \textbf{57.71}\scriptsize{$\pm$1.46} & &  22.25\scriptsize{$\pm$3.95} & &  \textbf{1.29}\\
SCL-EXP &  \underline{96.67}\scriptsize{$\pm$0.19} & &  77.16\scriptsize{$\pm$2.85} & &  86.56\scriptsize{$\pm$0.13} & &  81.80\scriptsize{$\pm$0.34} & &  34.43\scriptsize{$\pm$0.93} & &  52.25\scriptsize{$\pm$1.67} & &  \underline{22.42}\scriptsize{$\pm$2.46} & &  3.14\\
URE-NN &  94.07\scriptsize{$\pm$0.22} & &  75.61\scriptsize{$\pm$0.58} & &  84.54\scriptsize{$\pm$0.03} & &  58.05\scriptsize{$\pm$1.49} & &  21.66\scriptsize{$\pm$0.47} & &  40.24\scriptsize{$\pm$1.37} & &  14.76\scriptsize{$\pm$0.72} & &  7.43\\
URE-GA &  95.91\scriptsize{$\pm$0.12} & &  78.92\scriptsize{$\pm$0.80} & &  85.77\scriptsize{$\pm$0.15} & &  74.38\scriptsize{$\pm$1.49} & &  26.72\scriptsize{$\pm$0.49} & &  42.65\scriptsize{$\pm$2.90} & &  17.03\scriptsize{$\pm$1.59} & &  5.57\\
DM &  94.74\scriptsize{$\pm$0.21} & &  77.11\scriptsize{$\pm$1.60} & &  85.70\scriptsize{$\pm$0.27} & &  78.14\scriptsize{$\pm$0.52} & &  30.42\scriptsize{$\pm$1.40} & &  51.97\scriptsize{$\pm$2.77} & &  18.76\scriptsize{$\pm$2.53} & &  5.57\\
MCL-MAE &  96.53\scriptsize{$\pm$0.14} & &  74.37\scriptsize{$\pm$1.42} & &  86.35\scriptsize{$\pm$0.15} & &  65.15\scriptsize{$\pm$4.54} & &  11.93\scriptsize{$\pm$2.26} & &  34.04\scriptsize{$\pm$3.93} & &  11.47\scriptsize{$\pm$1.72} & &  7.00\\
MCL-EXP &  96.60\scriptsize{$\pm$0.10} & &  76.15\scriptsize{$\pm$2.72} & &  \underline{86.61}\scriptsize{$\pm$0.24} & &  80.43\scriptsize{$\pm$0.56} & &  33.98\scriptsize{$\pm$0.55} & &  56.20\scriptsize{$\pm$1.30} & &  21.74\scriptsize{$\pm$2.69} & &  3.86\\
MCL-LOG &  96.63\scriptsize{$\pm$0.05} & &  \underline{81.08}\scriptsize{$\pm$2.69} & &  86.57\scriptsize{$\pm$0.33} & &  \underline{82.36}\scriptsize{$\pm$0.58} & &  \underline{34.51}\scriptsize{$\pm$0.95} & &  \underline{56.35}\scriptsize{$\pm$1.75} & &  \textbf{23.03}\scriptsize{$\pm$2.54} & &  \underline{2.14}\\
\midrule
FWD &  \textbf{96.35}\scriptsize{$\pm$0.12} & &  \underline{80.35}\scriptsize{$\pm$1.89} & &  86.73\scriptsize{$\pm$0.27} & &  \textbf{81.77}\scriptsize{$\pm$0.63} & &  \underline{33.70}\scriptsize{$\pm$1.33} & &  \textbf{52.69}\scriptsize{$\pm$1.75} & &  \textbf{23.91}\scriptsize{$\pm$2.00} & &  \textbf{1.57}\\
URE-TNN &  93.94\scriptsize{$\pm$0.26} & &  76.00\scriptsize{$\pm$0.98} & &  85.16\scriptsize{$\pm$0.09} & &  59.66\scriptsize{$\pm$0.42} & &  21.49\scriptsize{$\pm$0.69} & &  36.96\scriptsize{$\pm$2.14} & &  14.87\scriptsize{$\pm$0.70} & &  5.71\\
URE-TGA &  95.99\scriptsize{$\pm$0.22} & &  79.35\scriptsize{$\pm$1.08} & &  85.92\scriptsize{$\pm$0.22} & &  74.09\scriptsize{$\pm$1.26} & &  25.63\scriptsize{$\pm$0.72} & &  44.23\scriptsize{$\pm$2.35} & &  17.13\scriptsize{$\pm$1.44} & &  4.00\\
CPE-I &  94.08\scriptsize{$\pm$0.42} & &  76.42\scriptsize{$\pm$0.69} & &  84.69\scriptsize{$\pm$0.18} & &  76.38\scriptsize{$\pm$0.88} & &  24.55\scriptsize{$\pm$0.73} & &  38.93\scriptsize{$\pm$0.97} & &  14.25\scriptsize{$\pm$1.21} & &  5.14\\
CPE-F &  \underline{96.34}\scriptsize{$\pm$0.12} & &  \textbf{80.36}\scriptsize{$\pm$1.89} & &  \underline{86.74}\scriptsize{$\pm$0.27} & &  81.68\scriptsize{$\pm$0.75} & &  \textbf{34.84}\scriptsize{$\pm$1.11} & &  \underline{52.64}\scriptsize{$\pm$1.67} & &  \underline{23.88}\scriptsize{$\pm$1.73} & &  \underline{1.86}\\
CPE-T &  96.28\scriptsize{$\pm$0.11} & &  79.02\scriptsize{$\pm$2.32} & &  \textbf{86.77}\scriptsize{$\pm$0.16} & &  \underline{81.69}\scriptsize{$\pm$0.58} & &  32.95\scriptsize{$\pm$1.14} & &  50.13\scriptsize{$\pm$2.29} & &  22.99\scriptsize{$\pm$1.37} & &  2.71\\
\bottomrule
\end{tabular}%
}
\end{table}

%% file: Tables/Multi-Simple-Tabular.tex
\begin{table}[t]
\centering
\caption{Comparison of test accuracies where instances have 3 CLs on Synthetic datasets and Real-World datasets.}
\label{tab:multi-simple-tabular}
\small
\resizebox{1\linewidth}{!}{%
\begin{tabular}{lccccccccccccccccccccccccccc}
\toprule
& \multicolumn{5}{c}{Synthetic Tabular Datasets} & \multicolumn{5}{c}{Real-World Datasets} \\
\cmidrule(lr){2-6} \cmidrule(lr){7-11}
 & {Yeast} & {Texture} & {Dermatology} & {Control} & Avg Rank & {CLCIFAR10} & {CLCIFAR20} & {CLMIN10} & {CLMIN20} & {Avg Rank} \\
\midrule
SCL-NL & \textbf{61.58}\scriptsize{$\pm$2.98} & \textbf{98.80}\scriptsize{$\pm$0.25} & \textbf{98.65}\scriptsize{$\pm$1.35} & \textbf{98.33}\scriptsize{$\pm$1.67} & \textbf{1.50} &  \textbf{47.30}\scriptsize{$\pm$0.50} & 8.59\scriptsize{$\pm$0.75} & 12.87\scriptsize{$\pm$2.33} & 6.87\scriptsize{$\pm$0.39} & 3.75\\
SCL-EXP & 60.40\scriptsize{$\pm$0.95} & 98.45\scriptsize{$\pm$0.50} & \textbf{98.65}\scriptsize{$\pm$1.35} & \underline{97.92}\scriptsize{$\pm$1.38} & \underline{2.88} &  \underline{47.12}\scriptsize{$\pm$0.91} & 9.74\scriptsize{$\pm$0.52} & 12.78\scriptsize{$\pm$1.42} & 7.10\scriptsize{$\pm$0.83} & 3.50\\
URE-NN & 51.68\scriptsize{$\pm$5.02} & 93.35\scriptsize{$\pm$0.37} & 85.81\scriptsize{$\pm$8.41} & 77.92\scriptsize{$\pm$11.51} & 6.50 &  40.74\scriptsize{$\pm$1.12} & \textbf{18.43}\scriptsize{$\pm$0.27} & \textbf{21.78}\scriptsize{$\pm$2.66} & \underline{7.16}\scriptsize{$\pm$0.70} & \underline{2.75}\\
URE-GA & 50.17\scriptsize{$\pm$3.33} & 85.86\scriptsize{$\pm$4.17} & 91.22\scriptsize{$\pm$6.99} & 73.75\scriptsize{$\pm$14.55} & 7.00 &  47.02\scriptsize{$\pm$0.45} & \underline{14.07}\scriptsize{$\pm$0.60} & \underline{20.83}\scriptsize{$\pm$2.88} & \textbf{7.71}\scriptsize{$\pm$1.24} & \textbf{2.00}\\
DM & \underline{61.24}\scriptsize{$\pm$1.53} & 97.70\scriptsize{$\pm$0.37} & \underline{97.97}\scriptsize{$\pm$2.24} & 96.25\scriptsize{$\pm$3.20} & 4.50 &  46.89\scriptsize{$\pm$0.32} & 9.11\scriptsize{$\pm$0.31} & 14.46\scriptsize{$\pm$2.22} & 6.84\scriptsize{$\pm$0.62} & 4.50\\
MCL-MAE & 33.22\scriptsize{$\pm$5.06} & 72.40\scriptsize{$\pm$8.37} & \textbf{98.65}\scriptsize{$\pm$1.35} & 70.42\scriptsize{$\pm$17.69} & 6.75 &  19.83\scriptsize{$\pm$2.91} & 7.88\scriptsize{$\pm$0.28} & 11.57\scriptsize{$\pm$0.80} & 6.64\scriptsize{$\pm$0.89} & 7.50\\
MCL-EXP & 59.90\scriptsize{$\pm$0.99} & 98.37\scriptsize{$\pm$0.35} & \textbf{98.65}\scriptsize{$\pm$1.35} & 97.50\scriptsize{$\pm$0.83} & 3.88 &  46.97\scriptsize{$\pm$1.23} & 8.42\scriptsize{$\pm$0.27} & 12.37\scriptsize{$\pm$1.61} & 6.62\scriptsize{$\pm$1.03} & 6.25\\
MCL-LOG & 60.40\scriptsize{$\pm$0.82} & \underline{98.67}\scriptsize{$\pm$0.43} & \textbf{98.65}\scriptsize{$\pm$1.35} & 97.50\scriptsize{$\pm$0.83} & 3.00 &  46.13\scriptsize{$\pm$0.57} & 8.57\scriptsize{$\pm$0.20} & 15.02\scriptsize{$\pm$2.25} & 6.55\scriptsize{$\pm$0.95} & 5.75\\
\midrule
FWD & \underline{61.74}\scriptsize{$\pm$2.07} & 98.32\scriptsize{$\pm$0.45} & \textbf{100.00}\scriptsize{$\pm$0.00} & \textbf{97.50}\scriptsize{$\pm$1.86} & 2.38 &  \textbf{52.48}\scriptsize{$\pm$0.63} & \textbf{24.56}\scriptsize{$\pm$0.95} & \underline{29.33}\scriptsize{$\pm$0.85} & \textbf{10.11}\scriptsize{$\pm$1.29} & \textbf{1.25}\\
URE-TNN & 44.30\scriptsize{$\pm$10.75} & 93.48\scriptsize{$\pm$1.19} & \underline{99.32}\scriptsize{$\pm$1.17} & 80.42\scriptsize{$\pm$6.60} & 4.75 &  35.60\scriptsize{$\pm$0.87} & 9.91\scriptsize{$\pm$2.98} & 17.98\scriptsize{$\pm$3.25} & 8.31\scriptsize{$\pm$0.59} & 5.25\\
URE-TGA & 41.11\scriptsize{$\pm$9.05} & 89.59\scriptsize{$\pm$0.54} & 97.97\scriptsize{$\pm$2.24} & 68.75\scriptsize{$\pm$5.70} & 5.75 &  45.08\scriptsize{$\pm$0.70} & 5.83\scriptsize{$\pm$1.36} & 17.95\scriptsize{$\pm$5.70} & 5.78\scriptsize{$\pm$0.14} & 5.75\\
CPE-I & 56.21\scriptsize{$\pm$4.28} & 97.39\scriptsize{$\pm$0.47} & 92.57\scriptsize{$\pm$2.95} & \underline{92.92}\scriptsize{$\pm$1.38} & 4.50 &  45.54\scriptsize{$\pm$2.83} & 20.41\scriptsize{$\pm$1.44} & 24.10\scriptsize{$\pm$2.26} & 8.48\scriptsize{$\pm$1.37} & 4.00\\
CPE-F & \textbf{62.08}\scriptsize{$\pm$1.81} & \textbf{98.57}\scriptsize{$\pm$0.54} & \textbf{100.00}\scriptsize{$\pm$0.00} & \textbf{97.50}\scriptsize{$\pm$1.86} & \textbf{1.62} &  \underline{51.74}\scriptsize{$\pm$0.98} & \underline{24.44}\scriptsize{$\pm$1.07} & \textbf{29.51}\scriptsize{$\pm$0.95} & 9.52\scriptsize{$\pm$1.71} & \underline{2.00}\\
CPE-T & 59.73\scriptsize{$\pm$2.89} & \underline{98.45}\scriptsize{$\pm$0.36} & \textbf{100.00}\scriptsize{$\pm$0.00} & \textbf{97.50}\scriptsize{$\pm$0.83} & \underline{2.00} &  49.79\scriptsize{$\pm$1.45} & 20.85\scriptsize{$\pm$0.52} & 27.97\scriptsize{$\pm$1.06} & \underline{9.70}\scriptsize{$\pm$1.13} & 2.75\\
\bottomrule
\end{tabular}%
}
\end{table}

%% file: Sections/Conclusion.tex
In this study, we introduce \texttt{libcll}, an open-source PyTorch library designed to advance research in complementary-label learning (CLL). The primary goal of \texttt{libcll} is to provide a standardized platform for evaluating CLL algorithms, addressing challenges in standardization, accessibility, and reproducibility. This library enables users to easily customize various components of the end-to-end CLL process, including data pre-processing utilities, implementations of CLL algorithms, and comprehensive metric evaluations that reflect realistic conditions. To demonstrate \texttt{libcll}'s flexible and modular design, we conduct diverse experiments encompassing multiple CLL algorithms, datasets ranging from synthetic to real-world, and various distribution assumptions.

Our experiments reveal that CPE and FWD are the most effective approaches for handling uniform, biased, and noisy complementary-label distributions. In cases where there is a substantial discrepancy between the known transition matrix and the actual distribution, such as when using an estimated transition matrix in real-world scenarios, we strongly recommend CPE. For multi-complementary-label learning in synthetic scenarios, SCL-NL algorithms are recommended. However, we note a limitation in handling deviations from a uniform distribution. For MCL from real-world datasets, our findings consistently show that CPE and FWD algorithms outperform other existing algorithms.


%% file: Sections/Limitation.tex
There are several ways to further enhance the comprehensiveness of the library. Currently, our strategy does not include other T-aware algorithms that may be competitive with CPE and FWD. Additionally, there is growing interest in leveraging complementary labels from similar instances to improve performance, a framework and functions for which are not yet supported in our library. In future work, we plan to integrate more recently published CLL algorithms and develop frameworks to accommodate more flexible labeling types in datasets.

\section{Broader impacts}
\label{sec:borader-impacts}
The library has the potential to advance algorithms for learning from complementary labels, enabling classifiers to be trained with limited information. However, this capability may increase the risk of compromising user privacy. We recommend that practitioners remain mindful of privacy concerns when using collected datasets and CLL algorithms.

%% file: reference.bbl
\begin{thebibliography}{10}

\bibitem{sugiyama2022machine}
Masashi Sugiyama, Han Bao, Takashi Ishida, Nan Lu, Tomoya Sakai, and Gang Niu.
\newblock {\em Machine learning from weak supervision: An empirical risk minimization approach}.
\newblock MIT Press, 2022.

\bibitem{zhou2018brief}
Zhi-Hua Zhou.
\newblock A brief introduction to weakly supervised learning.
\newblock {\em National science review}, 5(1):44--53, 2018.

\bibitem{ishida2017learning}
Takashi Ishida, Gang Niu, Weihua Hu, and Masashi Sugiyama.
\newblock Learning from complementary labels.
\newblock {\em Advances in neural information processing systems}, 30, 2017.

\bibitem{scl2020}
Yu-Ting Chou, Gang Niu, Hsuan-Tien Lin, and Masashi Sugiyama.
\newblock Unbiased risk estimators can mislead: A case study of learning with complementary labels, 2020.

\bibitem{mul-comp1}
Yuzhou Cao, Shuqi Liu, and Yitian Xu.
\newblock Multi-complementary and unlabeled learning for arbitrary losses and models.
\newblock {\em Pattern Recognition}, 124:108447, 2022.

\bibitem{mcl2020}
Lei Feng, Takuo Kaneko, Bo~Han, Gang Niu, Bo~An, and Masashi Sugiyama.
\newblock Learning with multiple complementary labels.
\newblock In {\em International Conference on Machine Learning}, pages 3072--3081. PMLR, 2020.

\bibitem{noisy_label}
Nagarajan Natarajan, Inderjit~S Dhillon, Pradeep~K Ravikumar, and Ambuj Tewari.
\newblock Learning with noisy labels.
\newblock In {\em Advances in Neural Information Processing Systems}. Curran Associates, Inc., 2013.

\bibitem{partial_labels}
Rong Jin and Zoubin Ghahramani.
\newblock Learning with multiple labels.
\newblock In {\em Advances in Neural Information Processing Systems}, volume~15, 2002.

\bibitem{fwd2018}
Xiyu Yu, Tongliang Liu, Mingming Gong, and Dacheng Tao.
\newblock Learning with biased complementary labels, 2018.

\bibitem{gao2021discriminative}
Yi~Gao and Min-Ling Zhang.
\newblock Discriminative complementary-label learning with weighted loss.
\newblock In {\em International Conference on Machine Learning}, pages 3587--3597. PMLR, 2021.

\bibitem{cpe2023}
Wei-I Lin and Hsuan-Tien Lin.
\newblock Reduction from complementary-label learning to probability estimates.
\newblock In {\em Proceedings of the Pacific-Asia Conference on Knowledge Discovery and Data Mining (PAKDD)}, May 2023.

\bibitem{clcifar2023}
Hsiu-Hsuan Wang, Wei-I Lin, and Hsuan-Tien Lin.
\newblock Clcifar: Cifar-derived benchmark datasets with human annotated complementary labels, 2023.

\bibitem{xu2019generativediscriminative}
Yanwu Xu, Mingming Gong, Junxiang Chen, Tongliang Liu, Kun Zhang, and Kayhan Batmanghelich.
\newblock Generative-discriminative complementary learning, 2019.

\bibitem{ComCo2023}
Haoran Jiang, Zhihao Sun, and Yingjie Tian.
\newblock Comco: Complementary supervised contrastive learning for complementary label learning.
\newblock {\em Neural Networks}, 169:44--56, 2024.

\bibitem{conu2023}
Wei Wang, Takashi Ishida, Yu-Jie Zhang, Gang Niu, and Masashi Sugiyama.
\newblock Learning with complementary labels revisited: A consistent approach via negative-unlabeled learning, 2023.

\bibitem{ishida2019complementarylabel}
Takashi Ishida, Gang Niu, Aditya~Krishna Menon, and Masashi Sugiyama.
\newblock Complementary-label learning for arbitrary losses and models, 2019.

\bibitem{ishiguro2022learning}
Hiroki Ishiguro, Takashi Ishida, and Masashi Sugiyama.
\newblock Learning from noisy complementary labels with robust loss functions.
\newblock {\em IEICE TRANSACTIONS on Information and Systems}, 105(2):364--376, 2022.

\bibitem{orderpreserving2023}
Shuqi Liu, Yuzhou Cao, Qiaozhen Zhang, Lei Feng, and Bo~An.
\newblock Consistent complementary-label learning via order-preserving losses.
\newblock In Francisco Ruiz, Jennifer Dy, and Jan-Willem van~de Meent, editors, {\em Proceedings of The 26th International Conference on Artificial Intelligence and Statistics}, volume 206 of {\em Proceedings of Machine Learning Research}, pages 8734--8748. PMLR, 25--27 Apr 2023.

\bibitem{wei2023classimbalanced}
Meng Wei, Yong Zhou, Zhongnian Li, and Xinzheng Xu.
\newblock Class-imbalanced complementary-label learning via weighted loss, 2023.

\bibitem{reg2021}
Deng-Bao Wang, Lei Feng, and Min-Ling Zhang.
\newblock Learning from complementary labels via partial-output consistency regularization.
\newblock In {\em IJCAI}, pages 3075--3081, 2021.

\bibitem{krizhevsky2009learning}
Alex Krizhevsky.
\newblock Learning multiple layers of features from tiny images.
\newblock pages 32--33, 2009.

\bibitem{gancl2023}
Jiabin Liu, Hanyuan Hang, Bo~Wang, Biao Li, Huadong Wang, Yingjie Tian, and Yong Shi.
\newblock Gan-cl: Generative adversarial networks for learning from complementary labels.
\newblock {\em IEEE Transactions on Cybernetics}, 53(1):236--247, 2023.

\bibitem{deng2022boosting}
Qinyi Deng, Yong Guo, Zhibang Yang, Haolin Pan, and Jian Chen.
\newblock Boosting semi-supervised learning with contrastive complementary labeling, 2022.

\end{thebibliography}
